\documentclass[runningheads]{llncs}

\usepackage{booktabs}
\usepackage{enumitem}
\usepackage{graphicx}
\usepackage{color}
\usepackage{fontenc}[T1]
\usepackage{inputenc}[utf8]

\usepackage{algorithm}
\usepackage{algorithmicx}
\usepackage{algpseudocode}

\newfloat{algorithm}{t}{lop}
\algnewcommand\algorithmicsymbols{\textbf{Symbols:}}
\algnewcommand\ASymbols{\item[\algorithmicsymbols]}
\algnewcommand\algorithmicinput{\textbf{Input:}}
\algnewcommand\Input{\item[\algorithmicinput]}
\algnewcommand\Output{\item[\algorithmicoutput]}

\begin{document}

\title{Unsupervised Concept Drift Detection based on~Parallel~Activations of~Neural~Network}
\titlerunning{Drift Detection on~Parallel~Activations of~Neural~Network}




\author{Joanna Komorniczak\orcidID{0000-0002-1393-3622} \\\and
Paweł Ksieniewicz\orcidID{0000-0001-9578-8395}}

\authorrunning{J. Komorniczak and P. Ksieniewicz}

\institute{Wrocław University of Science and Technology, \\ Department of Computer Systems and Networks}

\maketitle

\begin{abstract}
Practical applications of artificial intelligence increasingly often have to deal with the streaming properties of real data, which, considering the time factor, are subject to phenomena such as periodicity and more or less chaotic degeneration -- resulting directly in the \textit{concept drifts}. The modern concept drift detectors almost always assume immediate access to labels, which, due to their cost, limited availability, and possible delay, has been shown to be unrealistic. This work proposes an unsupervised \emph{Parallel Activations Drift Detector}, utilizing the outputs of an untrained neural network, presenting its key design elements, intuitions about processing properties, and a pool of computer experiments demonstrating its competitiveness with \emph{state-of-the-art} methods.

\end{abstract}

\section{Introduction}

The modern world is dominated by the mass production of data transmitted daily in petabytes of information traveling over the internet~\cite{feldmann2021year}. Artificial intelligence applications attempt to organize this chaotic reality from the level of dispersed information into knowledge, recently most often relying on \emph{semi-supervised} and \emph{unsupervised learning} methods, significantly reducing the need to rely on human experts in the labeling process~\cite{chen2020big}. However, default solutions of this type treat available data as static in time, often ignoring the phenomena of knowledge historicity and periodicity of concepts, thus striving to maximize efficiency within the full, huge volume of \emph{Big Data}~\cite{emmert2020introductory}. 

The field of \emph{machine learning} that focuses more on data velocity and considers the possibility of changing concepts between successive batches of incremental processing is \emph{Data Stream Processing}. One of the critical issues in this field is \emph{drift detection}, which involves identifying solutions that allow for effective signaling of significant changes in the concept. It should be noted, however, that the most common \emph{state-of-the-art} drift detectors most often assume full labeling of the data stream, which does not fit well into the increasingly dominant paradigm of \emph{semi-supervised} and \emph{unsupervised learning}. This shows the significant need to develop research on unsupervised drift detectors, which will potentially enable broad applications of \emph{Data Stream Processing} achievements in mainstream artificial intelligence research.

\subsection{Concept drift phenomenon}

Concept drift is taxonomically divided in terms of three main axes~\cite{agrahari2022concept}. According to the impact on recognition ability, drifts are divided into \textit{real} ones, the influence of which is visible when monitoring the quality of classification, and \textit{virtual}, which do not affect the decision boundary but may constitute the initial stage of \textit{real} changes. According to dynamics, distribution shift may occur at a single point in time (\textit{sudden} drift), or the transition can be spread over a longer period in \textit{gradual} and \textit{incremental} drifts. During \textit{incremental} drift, a temporary concept between the initial and target ones is observed, while in the case of \textit{gradual} change, objects from two consecutive concepts co-occur during the transition period. Finally, according to drift \textit{recurrence} -- a concept from the past may recur due to cyclical phenomena such as seasons or daily cycles. Additionally, the taxonomy considers drifts in which \textit{prior probability} of the classification problem changes~\cite{komorniczak2021prior}. Such drifts may affect recognition quality, showing falsely high accuracy values or a decrease in the quality when using metrics dedicated to imbalanced data, without a drift directly affecting the decision boundary.

\subsection{Related works}

According to the guidelines described by Domingos et al.~\cite{domingos2003general}, a~critical element of processing data streams is the mechanism for adapting to concept drifts. In the face of such changes in the data stream, two approaches are used: \textit{continuous rebuild} and \textit{triggered rebuild}~\cite{lindstrom2013drift}. In the case of the continuous rebuild, classifiers are trained throughout the entire processing period. In contrast, in the case of the triggered rebuild, specific determinants are used to indicate drifts, and only after the change is detected, the classifiers are updated to the current state of the posterior distribution of a stream. All approaches from the continuous rebuild strategy require almost immediate access to labels and use them to incrementally train the classifier. The factors used in the \textit{triggered rebuild} approach can be further divided into three categories: (\emph{a}) those monitoring the classification model, (\emph{b}) those monitoring the data, and (\emph{c}) those monitoring the output from the classification model~\cite{klinkenberg1998adaptive}. Similarly to the \textit{continuous rebuild}, the methods monitoring the classification output use labels for recognition quality assessment.

In the \textit{triggered rebuild} approach, the concept drift detectors are responsible for signaling the need to update the model. The first proposed drift detection methods took advantage of the fact that \textit{real} concept changes affect the recognition quality and monitored the frequency or distance of errors made by the classifier. Examples of such methods are \emph{Drift Detection Method} (\textsc{ddm})~\cite{gama2004learning} and \emph{Early Drift Detection Method} (\textsc{eddm})~\cite{baena2006early}. Subsequent detection methods used more complex mechanisms based on sliding windows in the \emph{Adaptive Windowing} (\textsc{adwin})~\cite{bifet2007learning} algorithm, pairs of classifiers in \emph{Paired Learners}~\cite{bach2008paired}, and ensembles of classifiers in \emph{Diversity for Dealing with Drifts} approach~\cite{minku2011ddd}. The main disadvantage of these solutions is a strong dependence on label access. It is worth mentioning here that \emph{implicit} supervised detectors have also been proposed, which do not directly rely on the classification quality to detect concept changes but use labels to analyze algorithm-independent properties of the data~\cite{hu2020no}. Regardless of how labels are used, the assumption of their almost immediate availability is not realistic due to limited access~\cite{sethi2015don}, their cost~\cite{liu2021online}, and possible time delay~\cite{grzenda2020delayed}. For those reasons, scientific interest in unsupervised drift recognition methods has increased in recent years~\cite{gemaque2020overview}. While some unsupervised methods monitoring the classification model will require access to labels, this is mainly to update the model after change detection~\cite{sethi2015don}. 

In accordance with the taxonomy described earlier, unsupervised drift detection methods will use two types of factors: those dependent on the classification model and those dependent on the data distribution itself. The data distribution is monitored in the \emph{Nearest Neighbor-based Density Variation Identification} (\textsc{nn}-\textsc{dvi})~\cite{liu2018accumulating} detector using the \textit{k-nearest neighbors} algorithm. Similarly, grid-based data distribution monitoring is proposed in the \emph{Grid Density based Clustering} (\textsc{gc}\oldstylenums{3})~\cite{sethi2016grid} approach. In the \emph{Centroid Distance Drift Detector} (\textsc{cddd})~\cite{klikowski2022concept} method, the distance between the centroids of subsequent batches is examined. This method can operate in both supervised and unsupervised modes. There are also methods based on the analysis of outlier observations, such as \emph{Fast and Accurate Anomaly Detection} (\textsc{faad})~\cite{li2019faad}, proposed mainly for the purpose of anomaly detection. In \emph{One-Class Drift Detector} (\textsc{ocdd})~\cite{gozuaccik2021concept}, a one-class classifier is used to examine the percentage of objects recognized as not belonging to the recognized concept. A similar strategy was used in \emph{Discriminative Drift Detector} (\textsc{d}\oldstylenums{3})~\cite{gozuaccik2019unsupervised}, where a discriminative classifier is used instead of a one-class classifier to explicitly recognize objects from the new concept from those from the previous one in variable-width windows.

Among the solutions based on the classifier properties, the \emph{Margin Density Drift Detection} (\textsc{md}\oldstylenums{3})~\cite{sethi2015don} method should be mentioned, in which the density of samples near the decision boundary of the \textsc{svm} classifier is examined. Similarly, in the \emph{Confidence Distribution Batch Detection} (\textsc{cdbd})~\cite{lindstrom2013drift} algorithm, the confidence of a classifier is monitored. Both of those approaches, despite their unsupervised detection, require access to labels in order to rebuild the monitored classifier in the case of a drift.

\subsection{Motivation and contribution}

In this work, we present a fully unsupervised \emph{Parallel Activations Drift Detector} (\textsc{padd}) method interpreting the activations of randomly initialized neural network (\textsc{nn}). Its overall detection mechanism shows some similarities to the \textsc{cdbd} detector -- which uses confidence in the outputs from the trained classifier (possibly \textsc{nn}) -- but without the requirement of label access to update the model in the event of drift. Similarly to \textsc{gc}\oldstylenums{3} method, \textsc{padd} employs the paradigm of original sample transformation into the condensed space, but on the contrary, it does not use a regular distribution grid, introducing non-uniform, tangled set of projections typical for the initial random state of a \textsc{nn}~\cite{narkhede2022review}.

The main contribution of this publication is to present a new drift detection method, operating on a fully unsupervised analysis of raw \textsc{nn} activations. The~work validates the overall quality of a proposition on synthetic data streams with various characteristics, comparing the proposed approach with unsupervised and supervised \emph{state-of-the-art} drift detection methods. Conducted experiments are publicly available to preserve the replicability of the research.

\section{\emph{Parallel Activations Drift Detector}}

This work presents an unsupervised drift detection method, operating purely on the output of a randomly initialized \textsc{nn}. The approach is described in the Algorithm~\ref{alg:pseudo}. 

\begin{algorithm}[!htb]
\scriptsize
\caption{Pseudocode of the \emph{Parallel-Activations Drift Detection}}
\label{alg:pseudo}
\algrenewcommand\algorithmicindent{0.86em}%
\vspace{-.125em}
\begin{algorithmic}[1]
\Input
\Statex $\mathcal{DS} = \{\mathcal{DS}_1, \mathcal{DS}_2, \ldots, \mathcal{DS}_k\}$ -- data stream,
\ASymbols
\Statex $\alpha$ -- significance level for statistical test
\Statex $\theta$ -- threshold for drift detection
\Statex $r$ -- statistical test replications performed for each NN output
\Statex $s$ -- number of samples drawn for statistical test
\Statex $\mathcal{C}$ -- stored activation values for all NN outputs since last drift
\Statex $\mathcal{NN}()$ -- forward pass from NN model with random weights and $e$ outputs
\Statex $\mathcal{S}()$ -- statistical test for sample subset comparison

\vspace{1em}

    \ForAll{$\mathcal{DS}_k \in \mathcal{DS}$}
    \State $c \gets \mathcal{NN}(\mathcal{DS}_k)$
    \Comment{Get ultimate network activations for current data chunk}
    \If{$\mathcal{C}$ is not empty}
        \State $a \gets 0$
        \Comment{Initialize counter as zero}
        \ForAll{$e_i \in e$}
            \ForAll{$r_i \in r$}
            \State $cc \gets $ random $s$ from $c[e_i]$
            \Comment{Randomly select samples from current and past certainty}
            \State $pc \gets $ random $s$ from $C[e_i]$
            \State $p \gets S(pc, cc)$
            \Comment{Compute p-value of statistical test}
            \If{$p < \alpha$}
                \State increment $a$
            \EndIf
            \EndFor
        \EndFor 
        \If{$a > \theta \times e \times r $}           
        \Comment{Drift detected}
            \State $\mathcal{C} \gets \emptyset$
            \State indicate drift in chunk $k$
        \EndIf       
    \EndIf
    \State store $c$ in $\mathcal{C}$
    \Comment{Store current activations for future comparison}     
    \EndFor
\end{algorithmic}
\vspace{-.125em}
\end{algorithm}

The method processes data streams divided into non-interlacing batches ($\mathcal{DS}_k \in \mathcal{DS}$). Drift detection is marked based on $r$ replications of statistical tests -- aiming to validate the null hypothesis stating the lack of significant difference between two groups of independent measurements -- comparing (\emph{a})~a~sample of size $s$ from the distribution of past and (\emph{b}) current activations at the all $e$ outputs of the \textsc{nn}. The initial -- and constant during the full processing -- random weights of \textsc{nn} are drawn from a normal distribution. As default settings, the normal distribution has an expected value of zero and a standard deviation of $0.1$. The statistical test used for distribution comparison is the Student's T-test for independent samples.

The critical parameters of the method are the significance level \textit{alpha} ($\alpha$) of the statistical test and the \textit{threshold} parameter ($\theta$), indicating the fraction of all tests that need to signal statistical independence of distributions to induce a~drift detection.

At the beginning of the processing of each batch, $c$ activations are calculated for samples from a given batch at all \textsc{nn} outputs. In the first chunk, the historical activations $\mathcal{C}$ are yet unknown, so the detection step is skipped due to the lack of reference data. The current activations $c$ are stored in the pool of historical outputs ($\mathcal{C}$). Otherwise, statistical tests are performed for individual network outputs. In lines \verb|5:14| of the pseudocode, $r$ replications of the statistical test are performed for each $e_i$ output of the network. Samples of size $s$ are drawn with replacement from historical activations for a given output $\mathcal{C}$ and for the current distribution $c$. If the statistical test shows a significant difference between the past and current distribution, the counter $a$ is incremented. The detection criterion is described in lines \verb|15:18| of the pseudocode. If the counter $a$ exceeds the required number of tests showing statistically significant (difference defined using $\theta$, the number of outputs $e$, and the number of test replications $r$) a concept drift in the current chunk is signaled. Such a signal implicates clearing the buffer of past supports ($\mathcal{C}$). For each batch, the current activations $c$ are saved to the historical data at the end of processing (line \verb|20|).

The Student's T-test shows a noticeably high sensitivity to the sample selection from the random variable provided to it. Therefore, the \textsc{padd} method stabilizes its verdict with replication of the measurement, which is possible thanks to a reliable buffer of historical activations. The invariance of the model weights, in turn, preserves the repeatability of the transformations performed by the \textsc{nn}, which should lead to results of low-dimensional embeddings to be statistically dependent in the absence of changes in the posterior distribution of the stream -- which we associate with both real and virtual drift phenomenon. Consequently, the proposed method is not built around the observation of the decision boundary -- as is the case with solutions basing detection on the evaluation of significant changes in the quality of processing -- but presents the potential to register general changes in the distribution occurring regardless of a given label bias.

\begin{figure}[!htb]
    \centering
    \includegraphics[width=\textwidth]{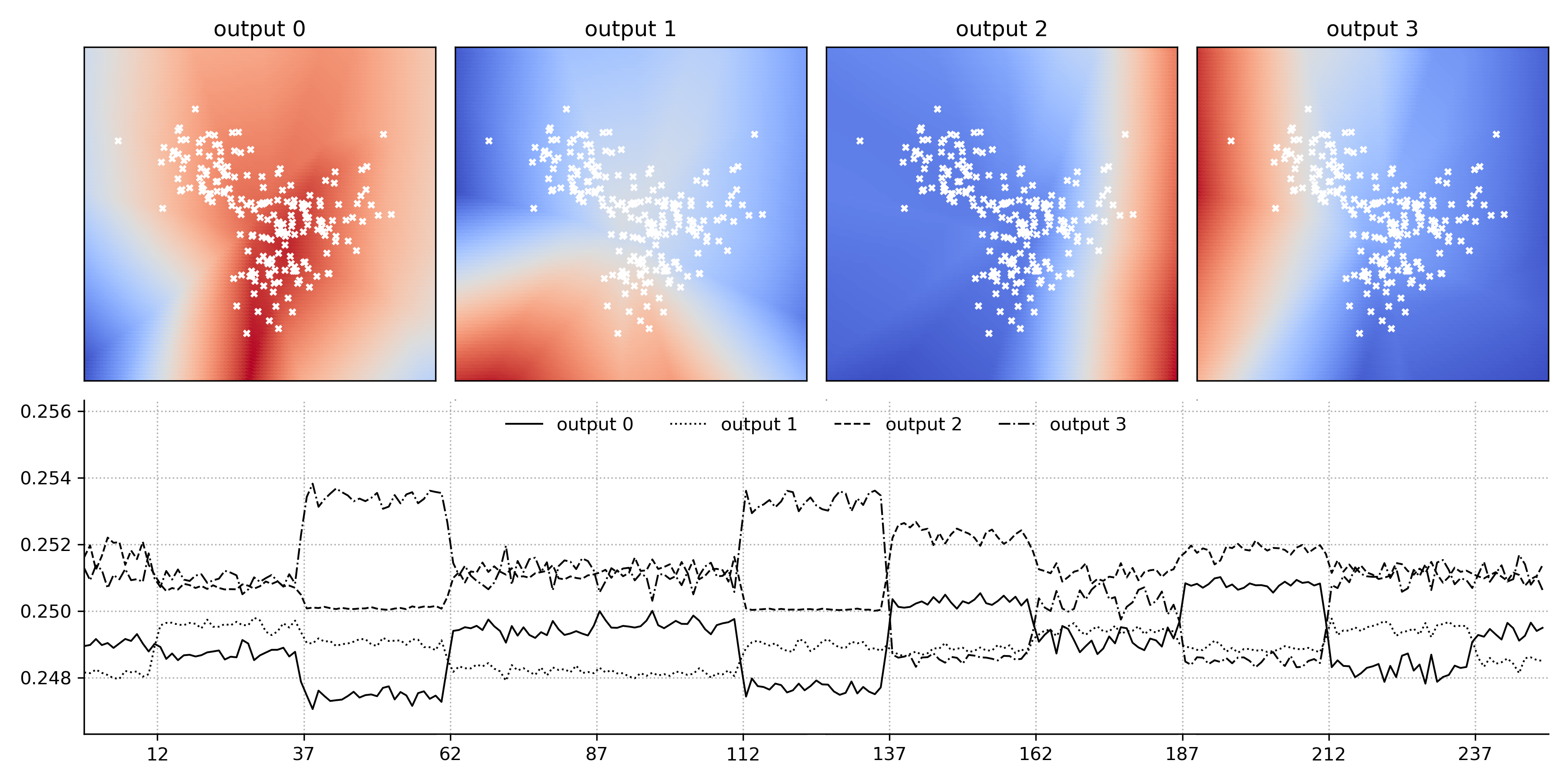}
    \caption{An example of 2-dimensional data (white points) presented in a context of ultimate layer activations of randomly initialized \textsc{nn} (top charts) and their mean activation (bottom chart) of four examined \textsc{nn} outputs during stream processing. Vibrant red areas in the top charts correspond to the high activation of a model, and vibrant blue to its low activation. The ticks on the horizontal axis of the bottom chart signal the moments of an abrupt concept change.}
    \label{fig:example}
\end{figure}

Figure~\ref{fig:example} shows the intuition behind the method operation on the exemplary stream with 250 data chunks and a final \textsc{nn} layer with four outputs. The first line presents the image (probing of a model with a mesh-grid covering two-dimensional feature space) of the \textsc{nn} output in the area sampled by data distribution. Red regions correspond to high activations, and blue to low activations. Random initialization of the weights causes diversified local landscapes in every dimension of the output space. In these areas, samples from one batch are marked with white markers. If drift occurs and the posterior distribution of a sampled data chunk changes, the structure of pseudo-supports in the recognized set will map this change within all or a part of the \textsc{nn} outputs. The second row of the graph presents the average activation values for batches in the data stream across all four outputs. The moments when drifts occur are clearly visible, additionally marked with chunk indices on the horizontal axis of the chart. In relation to the state of the network, some drifts will be easier to identify than others -- for example, the difference between the average output for the first drift (chunk 12) is less visible than for the next drift (chunk~37).

\section{Experiment design}

This chapter will describe the experimental protocol, the data streams used, and the goals of the experiment. The experiment and methods were implemented in the Python programming language, and the code is publicly available on the GitHub repository\footnote{\url{https://github.com/w4k2/padd}}.

\subsection{Data stream generation}

The experimental evaluation was performed on synthetic streams obtained using the generator from the \emph{stream-learn}~library~\cite{ksieniewicz2022stream}. The streams were described with various number of features and the number and type of drifts. The streams were processed in $250$ batches of $200$ samples, and each stream with a specific configuration was generated ten times with varying random states of a generator. 

The complete configuration of data streams is presented in Table~\ref{tab:config}. The~experiment compared the methods on the complete pool of $240$ data streams.

\begin{table}[]
    \centering
    \scriptsize
    \caption{Generator configuration for synthetic data streams}
    \setlength{\tabcolsep}{0.5em}
    {\renewcommand{\arraystretch}{1.4}
    \begin{tabular}{@{}lp{3.4cm}|lp{3.4cm}@{}}
    \toprule
    \textsc{parameter}          & \textsc{configuration} & \textsc{parameter}          & \textsc{configuration}    \\ \midrule
    Number of chunks   & 250                                & Drift frequency    & 3, 5, 10, 15 drifts\\
    Chunk size         & 200                                & Number of features & 30,  60,  90 (30\% informative) \\
    Drift dynamics     & sudden, gradual                    & Replications       & 10  \\\bottomrule 
    \end{tabular}}
    \label{tab:config}
\end{table}

The choice of synthetic data was dictated by the option to verify the operation of detection methods in various conditions and the possibility of replicating the stream generation to stabilize the results for statistical analysis. Additionally, only in the case of synthetic data the exact moments of drifts are known~\cite{lu2018learning}. This enables comparing changes signaled by the methods with the actual concept drifts.

\subsection{Measuring quality of drift detection}

The methods' detection quality was assessed using three \textit{drift detection error} measures~\cite{komorniczak2023complexity}, assessing the similarity of drifts occurring in the stream to those detected by a given method. The comparison protocol based on the classification quality has proven to show no relationship between the detection quality and the assessed accuracy of classification~\cite{bifet2017classifier}.

The three \textit{drift detection error} measures will evaluate detectors on three different criteria:
\begin{itemize}
    \item[D1] -- \textit{The average distance of each detection to the nearest drift}
    \item[] This measure will penalize the methods with numerous redundant drift detections, occurring far from the moments of actual concept changes. The average distance from detection to drift will increase in the case of hypersensitive methods. However, this measure does not consider the unrecognized concept changes -- hence, methods of low-sensitivity will not be penalized for lack of specific drift detection.
    
    \item[D2] -- \textit{The average distance of each drift to the nearest detection}
    \item [] This measure considers the closest detection of each actually occurring drift. In this measure, conversely to $D1$, the methods will be penalized for lack of drift signalization. However, the hypersensitivity of drift detectors will have no impact on this error measure -- as only the single detection closest to drift will be considered.
    
    \item[R] -- \textit{The adjusted ratio of the number of drifts to the number of detections}
    \item[] This error measure considers the number of drift that actually occur in the stream and the number of detections signaled by the method. The lowest possible error value of zero will be achieved for a method signaling the exact number of changes as a number of drifts. In case the method signals multiple redundant detections, the error will rise towards the value of one. If the method signals fewer detections due to low sensitivity, the error can rise to values above one. This measure, therefore, penalizes both too many and too few detections. Nevertheless, too little detection brings the error value higher, assuming that the lack of drift recognition is of greater significance during stream processing.
    
\end{itemize}
It is essential to note that measures can only be defined if the evaluated method signals any detection. Otherwise, the errors will be infinite, and statistical comparison will not be possible.

\subsection{Goals of the experiments}

\paragraph{\textbf{Hyperparameter selection}}

The first experiment aimed to select appropriate hyperparameters of the proposed method. Out of the five available hyperparameters only the two most critical were optimized: \textit{alpha} and \textit{threshold}. The remaining ones were fixed: 
\begin{itemize}
    \item the number of network outputs $e$ was 12, 
    \item the number of statistical test replications $r$ was 12, 
    \item the sample size $s$ was 50.
\end{itemize}

A neural network with a single hidden layer containing ten neurons and a~\emph{ReLU} activation function was used. 

For the \textit{alpha} parameter, 15 values from the range $0.03$ to $0.2$ were tested, and for the \textit{threshold} parameter, ten values from the range $0.1$ to $0.3$. The operation of the method with the indicated configurations was tested for the streams with ten drifts. The result of this experiment should indicate the range of values of these two parameters for which the method effectively recognizes drifts. Since both examined parameters indicate sensitivity to changes, their relationship should be visible.

\paragraph{\textbf{Comparison with the reference methods}}

The second experiment aimed to compare the proposed approach with reference methods. \emph{State-of-the-art} supervised and unsupervised detectors were selected. If possible, the implementation of methods provided by the authors was used or modified to allow processing streams in the form of data batches.

Table~\ref{tab:methods} presents all methods considered in the experiment, including the proposed \textsc{padd} method. The first column shows the acronym of the method, the second the category in the context of label access, the third the full name of the method, and a reference to the article introducing this approach. The last column describes the hyperparameterization of the method used in the experiment. 

\begin{table*}[!htb]
    \centering
    \scriptsize
    \caption{Method configuration for experimental evaluation}
    \setlength{\tabcolsep}{0.5em}
    {\renewcommand{\arraystretch}{1.4}
    \begin{tabular}{@{}p{1cm}p{1.6cm}p{2.5cm}p{6.125cm}@{}}
        \toprule
        \textsc{acronym} & \textsc{category} & \textsc{method name} & \textsc{selected hyperparameters} \\ \midrule
        
        \textsc{md}\oldstylenums{3} & Unsupervised with~label~request & \textit{Margin Density Drift Detection} \cite{sethi2015don} & \textit{threshold} parameter set depending on number of features: $0.15$ for 30 features, $0.1$ for 60 features and $0.08$ for 90 features \\\midrule
        
        \textsc{ocdd} & Unsupervised & \textit{One-Class Drift Detector} \cite{gozuaccik2021concept} &  \textit{percentage} parameter set depending on problem dimensionality: $0.75$ for streams with 30 features, $0.9$ for 60 features and $0.999$ in case of 90 features \\
        
        \textsc{cddd} & Unsupervised & \textit{Centroid Distance Drift Detector} \cite{klikowski2022concept} & \textit{sensitivity} parameter set depending on concept drift density: $0.2$ for streams with sparse changes (3,5) and $0.9$ for streams with dense changes (10,15) \\
        
       \textsc{padd}& Unsupervised & \textit{Parallel Activations Drift Detector} & configuration based on the results from the first experiment: 
       \textit{alpha} equal to $0.13$ for gradual drift and $0.07$ for sudden; \textit{threshold} of $0.26$ for gradual drift and $0.19$ for sudden;
       $r=12$; $e=12$; $s=50$
       \\\midrule
        
        \textsc{adwin} & Supervised &\textit{Adaptive Windowing} \cite{bifet2007learning} & default \textit{delta} of $0.002$, the base classifier used for error monitoring was Gaussian Naive Bayes\\
        
        \textsc{ddm} & Supervised & \textit{Drift Detection Method} \cite{gama2004learning} & default detection \textit{threshold} of $3$, the base classifier used for error monitoring was Gaussian Naive Bayes\\
        
        \textsc{eddm} & Supervised & \textit{Early Drift Detection Method} \cite{baena2006early} & default \textit{beta} of $0.9$, the base classifier used for error monitoring was Gaussian Naive Bayes\\\bottomrule
    \end{tabular}}
    \label{tab:methods}
\end{table*}

The default parameters were selected for supervised methods, consistent with the implementation in the \emph{scikit-multiflow} library~\cite{skmultiflow}. For the remaining approaches, the hyperparameters were manually selected according to the articles introducing the methods or altered to effectively process the evaluated types of streams.

\section{Experimental evaluation}

This section will present and analyze the results of performed experiments.

\subsection{Hyperparameter selection}

The first experiment aimed to select the appropriate method hyperparameters. The analysis focused on two variables describing the method sensitivity -- \emph{alpha} value for statistical test significance and \emph{threshold} value for method integration. 

\begin{figure}[!htb]
    \centering
    \includegraphics[width=\textwidth]{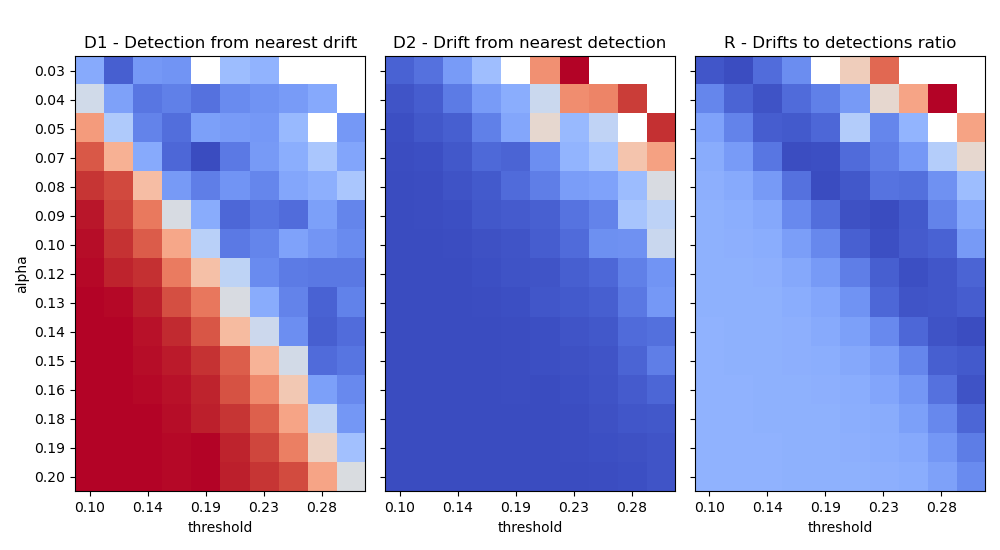}
    \caption{Drift detection error measures for a single 30-dimensional data stream with sudden concept drifts, depending on the values of two critical hyperparameters -- \emph{alpha} and \emph{threshold}, describing the method sensitivity. Red cells indicate high errors, while blue cells -- the low error.}
    \label{fig:c2d_e0_individual}
\end{figure}

The results for three drift detection error measures and a single stream described by 30 features and characterized by sudden concept drifts are presented in Figure~\ref{fig:c2d_e0_individual}. The heatmaps present the $D1$, $D2$ and $R$ errors, respecively. Blue cells describe low error values, while red ones indicate a high error. The cells that are left blank indicate that the error was not possible to be measured due to the lack of detections. Each heatmap's vertical axis describes the \emph{alpha} parameter values, and the horizontal axis describes the \emph{threshold} parameter. 

The lower left area of each heatmap -- for the low \emph{threshold} values and high \emph{alpha} -- describes a high sensitivity to changes present in the stream and is related to numerous redundant detections, visible in high $D1$ error and significant $R$~error. Meanwhile, the upper right area describes the low sensitivity of a method -- where all error values, including $D2$, are high. For the very low sensitivity of the method, no detections will be returned -- hence the visibility of white cells. The best configuration of the method will be parameter values between these two extremes, where all three drift detection measures return low error values.

\begin{figure}[!htb]
    \centering
    \includegraphics[width=\textwidth]{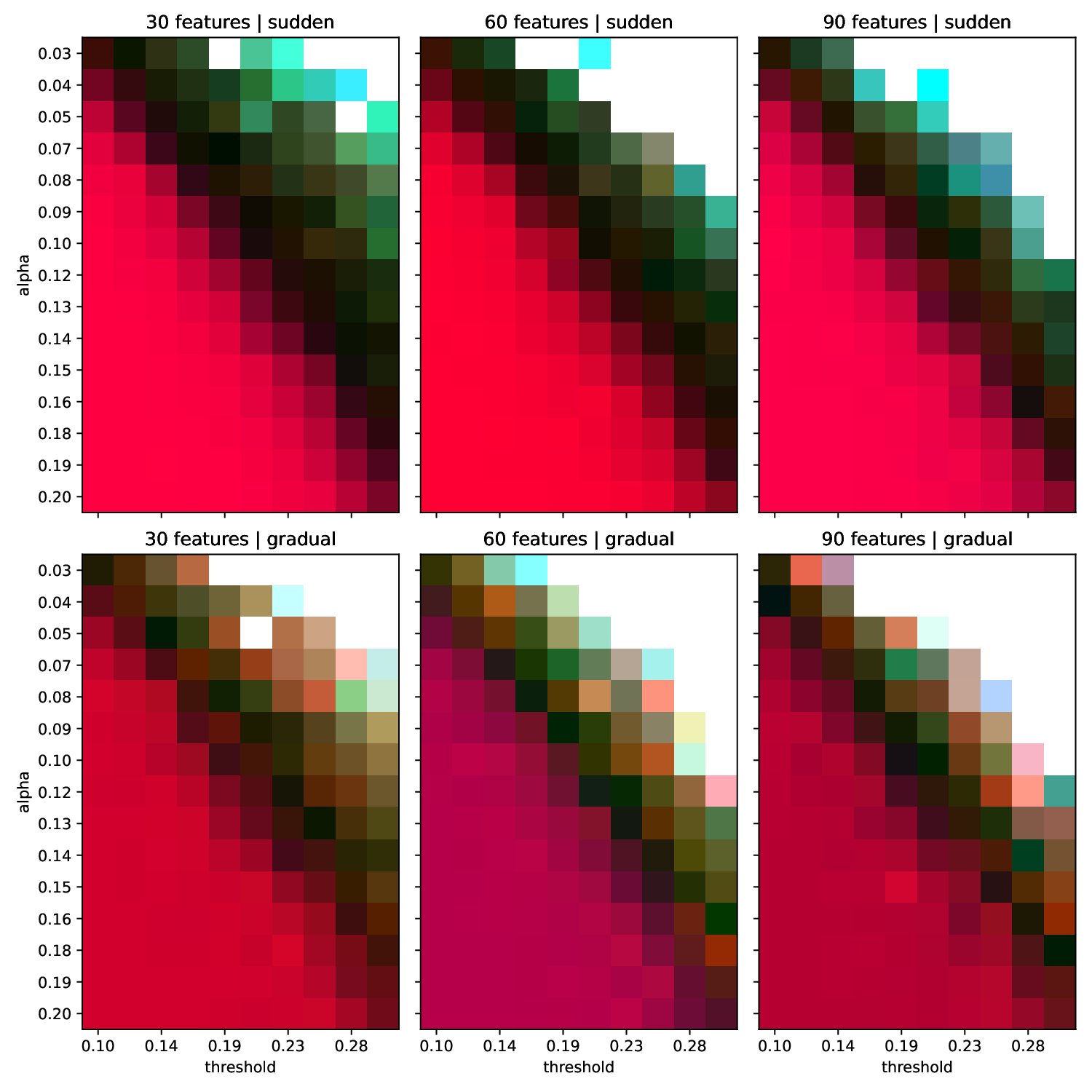}
    \caption{The combination of drift detection error measures, normalized to a range 0-1 in each of three measures, presented as an RGB image. Black cells indicate the lowest error across all three measures.}
    \label{fig:c2d_e0}
\end{figure}

Experiment results for all streams analyzed in the first experiment are presented in Figure~\ref{fig:c2d_e0}. The color heatmaps show the combination of three \emph{drift detection error} measures after their normalization. The red color channel corresponds to $D1$ measure, green to $D2$, and blue to $R$ error. The results for streams with sudden drifts are presented in the first row and for gradual concept change in the second. The columns present various dimensionalities of the data -- 30, 60, and 90 features, respectively. After such a color combination of error values, the lowest errors in all three criteria will be marked by colors close to black. 

It is worth noting that the method's effectiveness is highly dependent on the selection of these two parameters, and the lack of precise selection may result in excessive detection or failure to recognize drifts. Ultimately, the following parameter combinations were selected for the comparison experiment: for \emph{gradual} drifts -- \emph{alpha} equal to 0.13 and \emph{threshold} equal to 0.26; for \emph{sudden} drifts -- \emph{alpha} equal to 0.07 and \emph{threshold} of 0.19.

\subsection{\textbf{Comparison with reference methods}}

The second experiment compared the performance of the proposed approach with reference drift detectors. 

\begin{figure}[!htb]
    \centering
    \includegraphics[width=\textwidth]{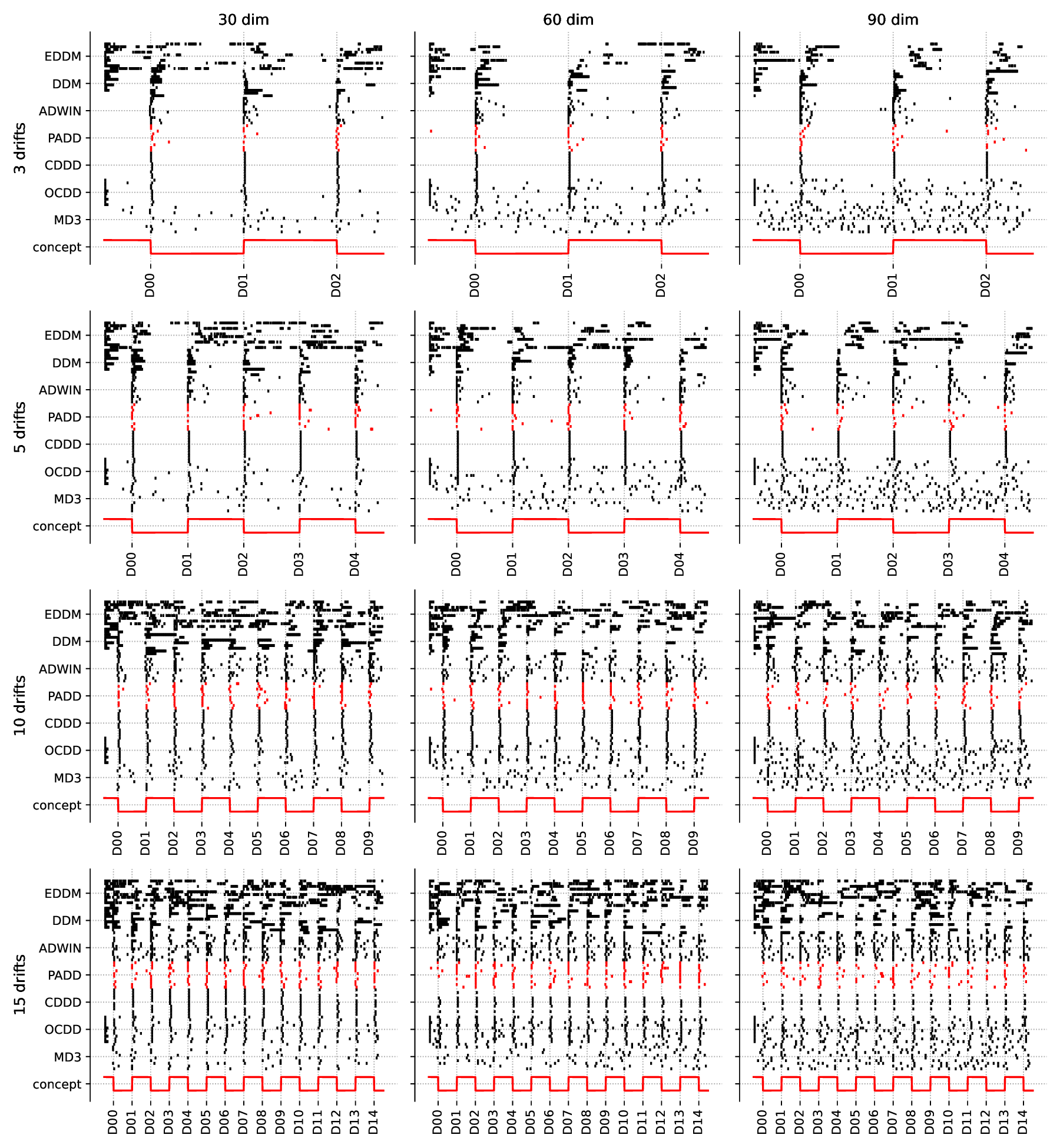}
    \caption{Detection moments of all evaluated methods for streams with \textit{sudden} concept drifts. The actual drifts are marked with ticks on the horizontal axis and particular methods's detections in ten stream replications with black or red (in the case of the proposed approach) points. The ideal drift detection would result in vertical lines overlapping with ticks on x-axis.}
    \label{fig:padd_exp1-sudden}
\end{figure}

The results for streams with ten drifts are shown in Figure~\ref{fig:padd_exp1-sudden}. The columns indicate the results for different numbers of features -- from 30 in the first column to 90 in the last one -- and the rows for different numbers of drifts present in the stream -- from 3 drifts in the first row to 15 drifts in the last row. On the horizontal axis of each graph, successive chunks of the data stream are visible, while the central moments of the actual drift are marked with ticks on the x-axis and a grid. Each detection is marked with a single point. For emphasis, the proposed approach is shown in red. The consecutive lines show the results from subsequent replications for a given detector. The last row shows the concept drift dynamics.

These graphs allow to visually assess the detection quality. The ideal result would be for the method to flag a single change at the center point of each drift occurrence across all replications -- this would result in vertical lines overlapping with the drift markings on the horizontal axis. 

\begin{figure}[!htb]
    \centering
    \includegraphics[width=\textwidth]{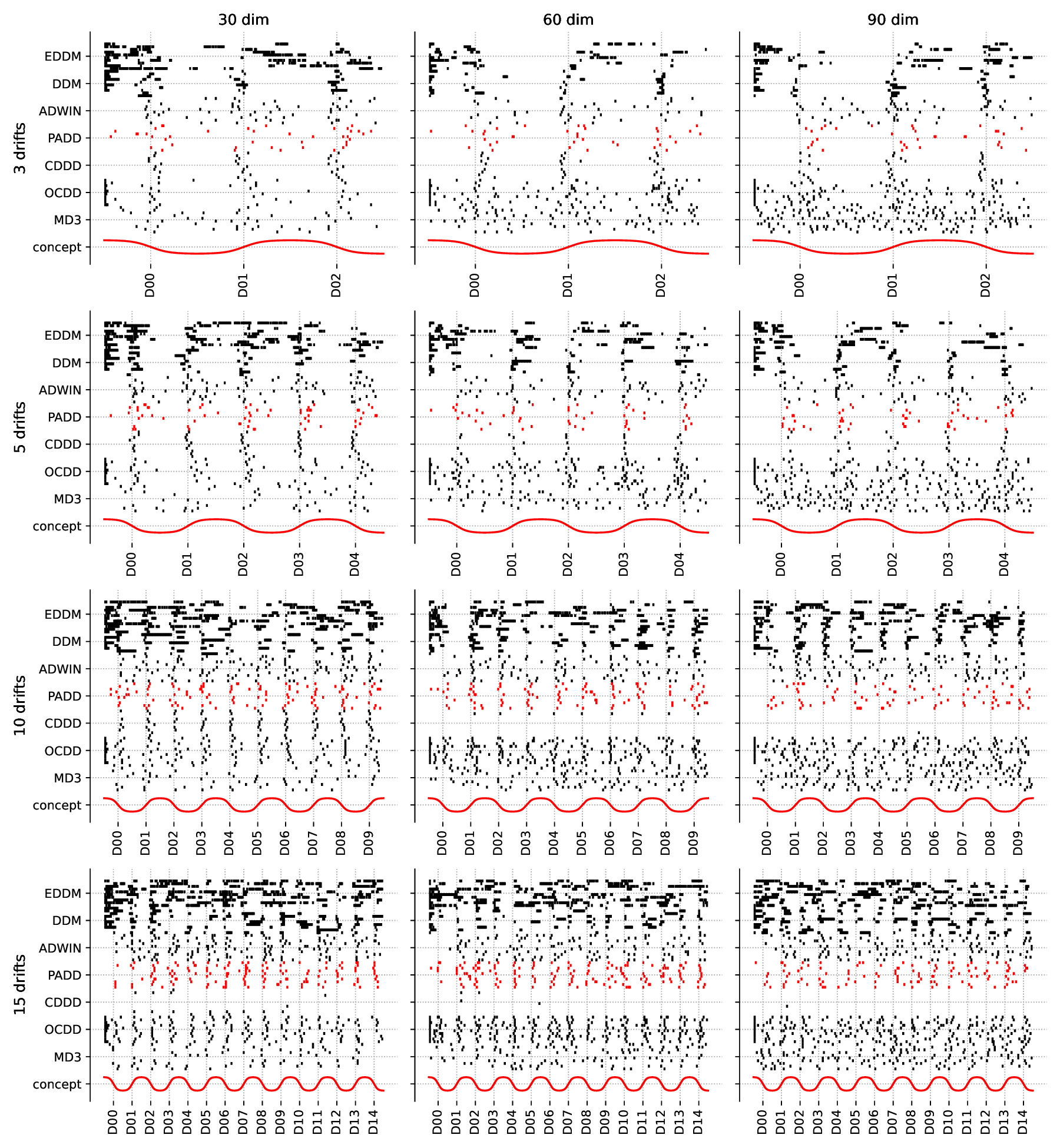}
    \caption{Detection moments of all evaluated methods for streams with \textit{gradual} concept drifts. The ticks on x-axis indicate the central point of each concept drift.}
    \label{fig:padd_exp1-gradual}
\end{figure}

Figure~\ref{fig:padd_exp1-gradual} shows the results for streams with gradual concept drift similarly. It can be seen that detections for all methods are more dispersed in the case of gradual drifts -- this is either due to the detection in an early or later phase of drift or due to multiple signaling of the same drift stretched over time in the case of high sensitivity of the method. 

\begin{figure*}[!htb]
    \centering
    \includegraphics[width=\textwidth]{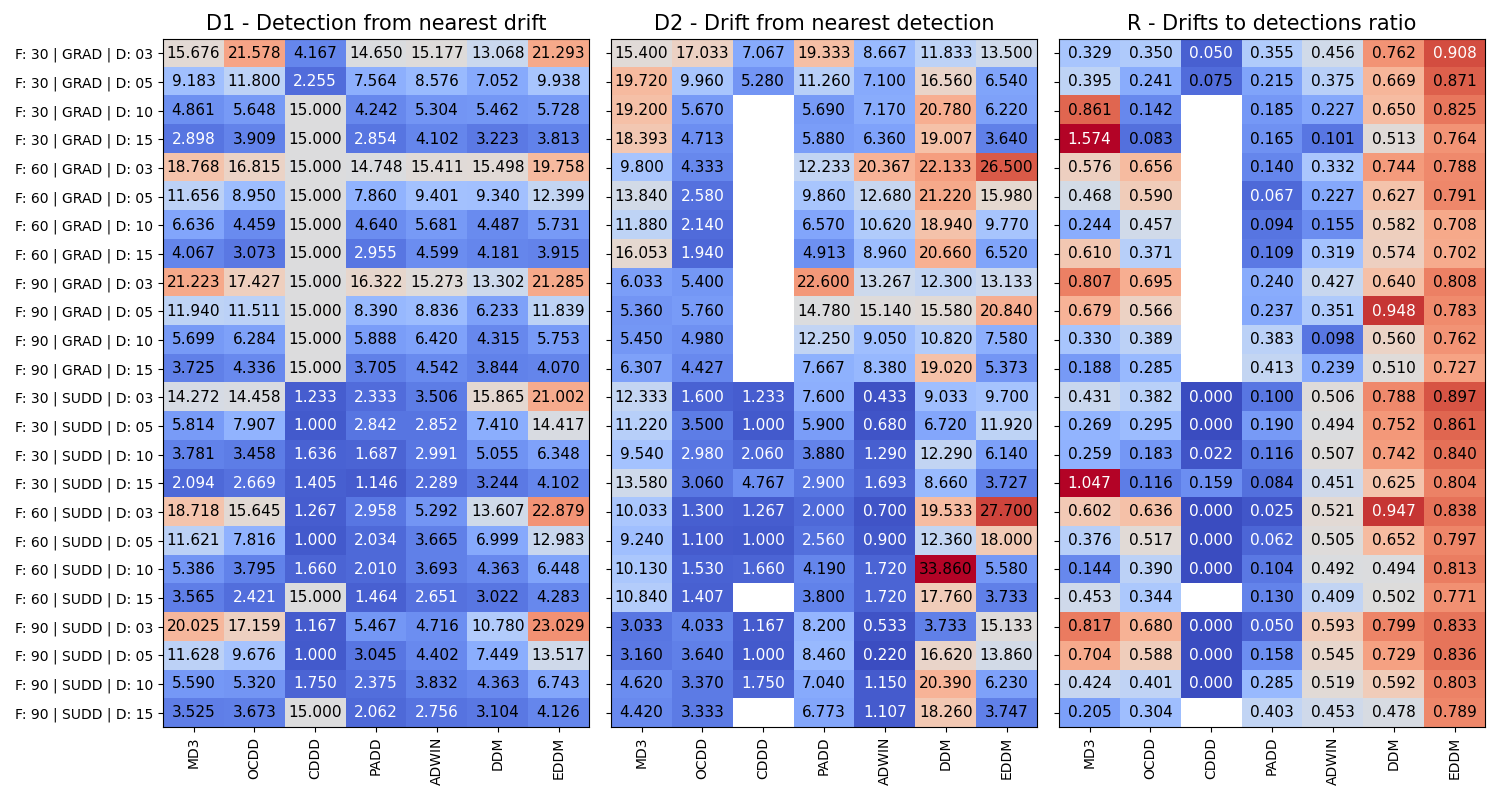}
    \caption{Drift detection error measures: D1 (left), D2 (center), and R (right) for all evaluated detection methods (horizontal axis) and all evaluated data streams (vertical axis). High error values are marked in red color; low values in blue.}
    \label{fig:exp1-comp}
\end{figure*}

Looking at the color-coded tables in the form of heat maps, shown in Figure~\ref{fig:exp1-comp}, allows for a more general interpretation. The low error values are marked in blue and high in red. Cases for which it was impossible to calculate errors -- due to lack of detection -- are left as blank cells. The columns of the figure show the results for the three drift detection error measures -- $D1$, $D2$, and $R$, respectively. In each heat map, the results for individual methods are presented in columns, and the results for all tested types of streams are presented in rows, averaged over ten replications.

In the first error, \textit{the average distance of each detection to the nearest drift} ($D1$), the average distance to the nearest drift is measured for each detection. This means that methods that signal many distant drifts from the moment of their actual occurrence, such as \textsc{ddm}, \textsc{eddm}, and \textsc{md}\oldstylenums{3}, will achieve higher error values. It is also visible that for streams where drifts are less frequent (3 and 5 in the entire stream), these errors' values will be higher due to the relatively greater distance for incorrect detections. 

In the second measure, \textit{the average distance of each drift to the nearest detection} ($D2$), a single nearest detection is searched for each drift. The $D2$ measure will penalize the methods that fail to recognize the actually occurring drifts or do so with a delay. The highest errors in this criterion are demonstrated by the \textsc{ddm} and \textsc{md}\oldstylenums{3} methods, which sometimes signaled drifts with a delay or did not detect all changes. 

The last measure, \textit{the adjusted ratio of the number of drifts to the number of detections} ($R$), looks only at the actual number of drifts compared to the number of changes signaled by the method.  In the case of too many detections, the measure will return values close to (but not exceeding) one. However, when the number of signaled changes is lower than the number of drifts, the measure will return values exceeding one. High error values are visible for the \textsc{ddm} and \textsc{eddm} methods, but they do not exceed one, hence the methods were penalized for too many detections. However, the \textsc{md}\oldstylenums{3} method obtained measure values exceeding one, which indicates too few detections on average, compared to the number of drifts. It is worth paying attention to the results for the \textsc{cddd} method, which are exceptionally low in all streams for which the measure could be determined, but its problem is the tendency to not signal frequent and gradual drifts.

\begin{figure}[!htb]
    \centering
    \begin{tabular}{ccc}
        ($D1$)&
        ($D2$) &
        ($R$)\\

        \includegraphics[trim={1cm 0 1cm 0 },clip,width=0.33\textwidth]{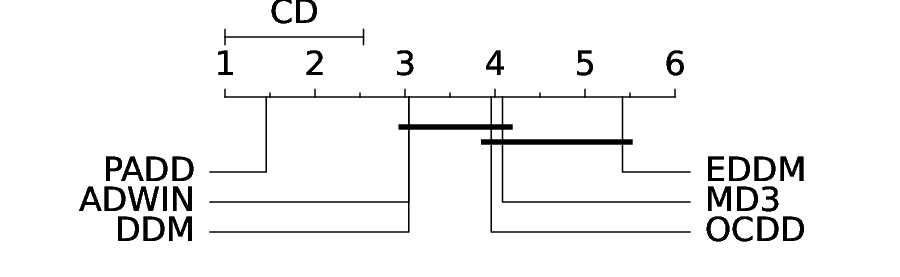} &
        \includegraphics[trim={1cm 0 1cm 0 },clip,width=0.33\textwidth]{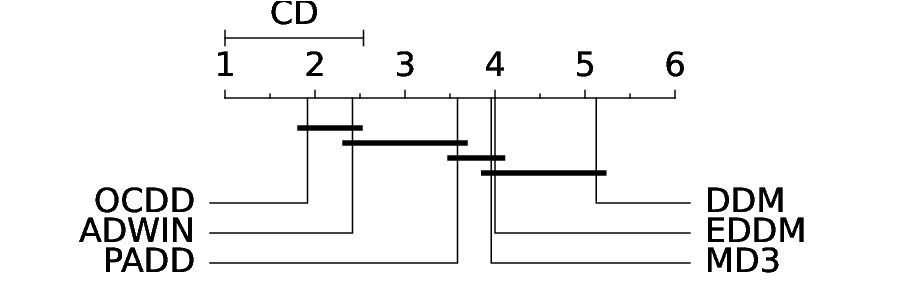} &
        \includegraphics[trim={1cm 0 1cm 0 },clip,width=0.33\textwidth]{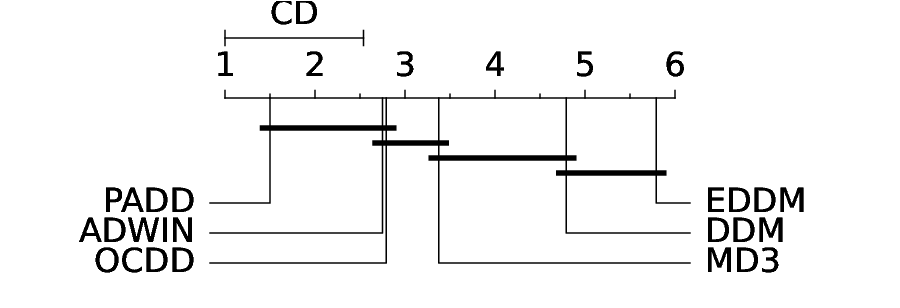} \\
    \end{tabular}
    
    \caption{Critical Difference diagrams for three drift detection error measures: D1 (left), D2 (center), and R (right). The \textsc{cddd} method was not taken into account in the statistical evaluation due to the lack of possibility to calculate \textit{drift detection measures} in certain data streams.}
    \label{fig:cd}
\end{figure}

The results obtained from the second experiment were also subjected to post-hoc \emph{Nemenyi} test based on \emph{Wilcoxon Signed Rank} across all three errors. The results for $D1$, $D2$, and $R$ errors are presented in the Figure~\ref{fig:cd}. The overall statistical analysis did not consider the results of the \textsc{cddd} detector due to the inability to calculate measures in the absence of detection by the method, which took place in 12 out of 24 calculations.

The results show that in the case of $D1$ error, the presented method is significantly the best lone leader of comparison. For the $R$ error, the presented solution -- while having the best ranking value -- is statistically significantly co-dependent to the \textsc{adwin} and \textsc{occd}. For the $D2$, the best results were obtained by the \textsc{ocdd} method, and the \textsc{padd} results are statistically dependent on the second method in terms of quality, the \textsc{adwin} method with the minimal difference in ranks.

It is worth emphasizing that those criteria should not be used independently to evaluate methods, and it is the juxtaposition of all three that describes the proper and effective method operation.

\section{Conclusions and future works}

This work proposed an unsupervised \emph{Parallel Activations Drift Detector} that utilizes an untrained neural network to recognize significant changes in the posterior distribution of the data stream to signal concept drift according to the result of a statistical test stabilized by a pool of replications. The conducted experimental evaluation allowed to demonstrate that the proposed \textsc{padd} method states a valuable tool in the context of reference methods. This enriches the pool of available drift detection methods, introducing an algorithm with high reliability in a proper configuration.

As part of future work, it is planned to expand the hyperparameter calibration study -- considering the introduction of a non-parametric version of the method interpreting the area under the curve of the threshold function of a statistical test -- and a broader review of possible validators to replace the simple Student's T-test. An interesting area for developing the analysis would also be to conduct research on the dependencies between the outputs of the \textsc{nn} used, allowing for the naivety of their independence to be abandoned.

\subsection*{Acknowledgements}
This work was supported by the statutory funds of the Department~of~Computer~Systems~and~Networks, Wroclaw~University~of~Science~and~Technology.

\bibliographystyle{splncs04}
\bibliography{biblio}

\end{document}